\documentclass[cameraready]{Interspeech}
\title{Phoneme- vs.\ Character-Level Targets and Selective State-Space Models for Intracortical Brain-to-Text}

\author[affiliation={1}]{Lucas}{Zamora Vera}
\author[affiliation={2,3,1}, orcid=0000-0002-5531-8994]{Jose A.}{Gonzalez-Lopez}

\address{
   $^1$ Universitat Oberta de Catalunya (UOC), Spain \\
   $^2$ Dpt.\ of Signal Theory, Telematics and Communications,  University of Granada,  Spain \\
   $^3$ Research Centre for Information and Communication Technologies (CITIC-UGR),  University of Granada,  Spain
}

\email{lzamorav@uoc.edu, joseangl@ugr.es}


\usepackage{comment}
\usepackage{tikz}
\usetikzlibrary{arrows.meta, positioning}
\begin{document}

\maketitle

% the abstract here must exactly match the abstract entered into the paper submission system
\begin{abstract}
    % 1000 characters. ASCII characters only. No citations.
    State-of-the-art intracortical brain-to-text systems pair a neural-sequence phone decoder with an external language model. Two design axes remain underexplored: whether selective state-space models (Mamba) improve on recurrent decoders, and how the output target (phonetic vs.\ character) interacts with that choice. On the public Brain-to-Text '25 benchmark, we study a controlled 2x2 grid (GRU vs.\ hybrid Mamba decoder; phonetic vs.\ character targets) trained with a CTC objective under one reproducible protocol. The recurrent baseline remains strongest: the best phonetic GRU reaches 12.62\% PER and 21.19\% WER, while the best textual GRU after LM rescoring reaches 13.39\% CER and 26.28\% WER. The Mamba hybrid is competitive but does not surpass it. Ablations isolate architectural contributions, and error analysis shows representation-dependent failures: articulatory-like phoneme confusions vs.\ lexical and word-boundary errors.

\end{abstract}

\section{Introduction}

People with amyotrophic lateral sclerosis (ALS), brainstem stroke or locked-in syndrome can lose intelligible speech while retaining cognition, and the recovery of a reliable communication channel substantially improves autonomy and social participation~\cite{silva2024speech, wolpaw2002bci, chang2024restoring}. Brain-computer interfaces (BCIs) aim to restore this channel by translating neural activity into communicative output~\cite{wolpaw2002bci}, a paradigm first established for motor and cursor control in people with paralysis~\cite{hochberg2012reach, pandarinath2017high, vansteensel2016fully}. Speech decoding has been pursued across recording modalities: non-invasive recordings support coarse perception decoding~\cite{tang2023semantic, defossez2023decoding}, electrocorticography (ECoG) enables phrase- and sentence-level decoding and synthesis~\cite{herff2015braintotext, makin2020machine, anumanchipalli2019synthesis, angrick2019synthesis, martin2014decoding}, and intracortical microelectrode arrays offer the highest spatiotemporal resolution, driving the most accurate recent speech decoders~\cite{silva2024speech, willett2023high, card2024accurate}. The neurofunctional basis for this is that ventral sensorimotor cortex retains rich articulatory information even when motor output is lost~\cite{silva2024speech, hickok2007cortical, pulvermuller2018neural}.

A \textit{brain-to-text} system casts speech decoding as a sequence-to-sequence problem with no explicit alignment between the multichannel neural signal and the target symbol sequence, which is naturally handled by the Connectionist Temporal Classification (CTC) objective~\cite{graves2006ctc}. The dominant paradigm is two-stage: a recurrent decoder maps neural activity to sub-lexical units (typically phonemes), and a language model (LM) reconstructs the most plausible sentence~\cite{willett2023high, card2024accurate, silva2024speech, brumberg2010bci}. Landmark intracortical results, including high-rate handwriting decoding~\cite{willett2021handwriting}, real-time decoding in anarthria~\cite{moses2021neuroprosthesis}, a large-vocabulary speech neuroprosthesis~\cite{willett2023high}, a high-performance avatar-and-speech system~\cite{metzger2023avatar}, generalizable spelling~\cite{metzger2022spelling}, instantaneous voice synthesis~\cite{wairagkar2025voice}, and a rapidly calibrating system reaching sub-5\% word error rates over extended use~\cite{card2024accurate}, established the clinical relevance of this line of work, approaching the accuracy of modern automatic speech recognition (ASR) systems~\cite{baevski2020wav2vec, radford2023whisper}. The recent Brain-to-Text '24 benchmark reported that the largest gains in decoding performance came from better training, ensembling, and LM rescoring rather than from replacing the recurrent decoding backbone, with Transformers showing no clear advantage~\cite{willett2024benchmark, li2024context}.

Two design choices, however, remain comparatively under-studied. First, selective state-space models (SSMs), specifically Mamba~\cite{gu2023mamba}, offer a linear-time recurrent formulation with input-dependent selectivity that is attractive for long, noisy neural sequences. SSMs have shown promise in ASR, both as standalone speech encoders~\cite{miyazaki2024mamba, jiang2024slytherin} and in convolution-augmented forms~\cite{hou2024conmamba, zevallos2025mamba} that echo the success of attention-based sequence models~\cite{vaswani2017attention, chan2016listen}, the Conformer~\cite{gulati2020conformer} and its efficient variants~\cite{burchi2021efficient, peng2022branchformer}. Whether they help for intracortical decoding is an open question. Second, most systems commit to a phonetic target, yet a direct character-level target removes the dependency on a pronunciation lexicon and brings the decoder closer to the final output. A recent end-to-end character-level Conformer study on the same dataset showed that meaningful direct character decoding is possible without an external language model, but also found that dominant errors arise from incorrect word-boundary segmentation~\cite{khanday2026endtoend}. However, a controlled comparison of the \textit{target representation} crossed with the \textit{decoder architecture} remains missing.

This paper asks: \textit{do selective state-space models improve on recurrent decoders for intracortical brain-to-text, and how does the output target (phoneme vs.\ character) interact with that choice?} We make three contributions: (i) a controlled $2\times2$ study (GRU vs.\ a hybrid ConvMambaGRU; phonetic vs.\ textual targets) under one reproducible CTC pipeline on the public Brain-to-Text '25 benchmark; (ii) to our knowledge, the first adaptation of a ConvMamba-style backbone to intracortical brain-to-text, with ablations isolating its convolutional front-end and recurrent refinement stage; and (iii) a representation-aware error analysis contrasting phoneme-level and character-level failure modes. Our finding is that the recurrent baseline remains the strongest decoder and the phonetic two-stage variant gives the lowest WER; the SSM hybrid is competitive but does not surpass it, consistent with the data-limited regime of single-participant intracortical decoding~\cite{willett2024benchmark}.

% Related work is folded into this Introduction per the agreed structure; keep Intro + abstract within ~1 page.

\section{Methods}

\subsection{General procedure}
\label{sec:procedure}
\begin{figure}[t]
    \centering
    \includegraphics[width=\linewidth]{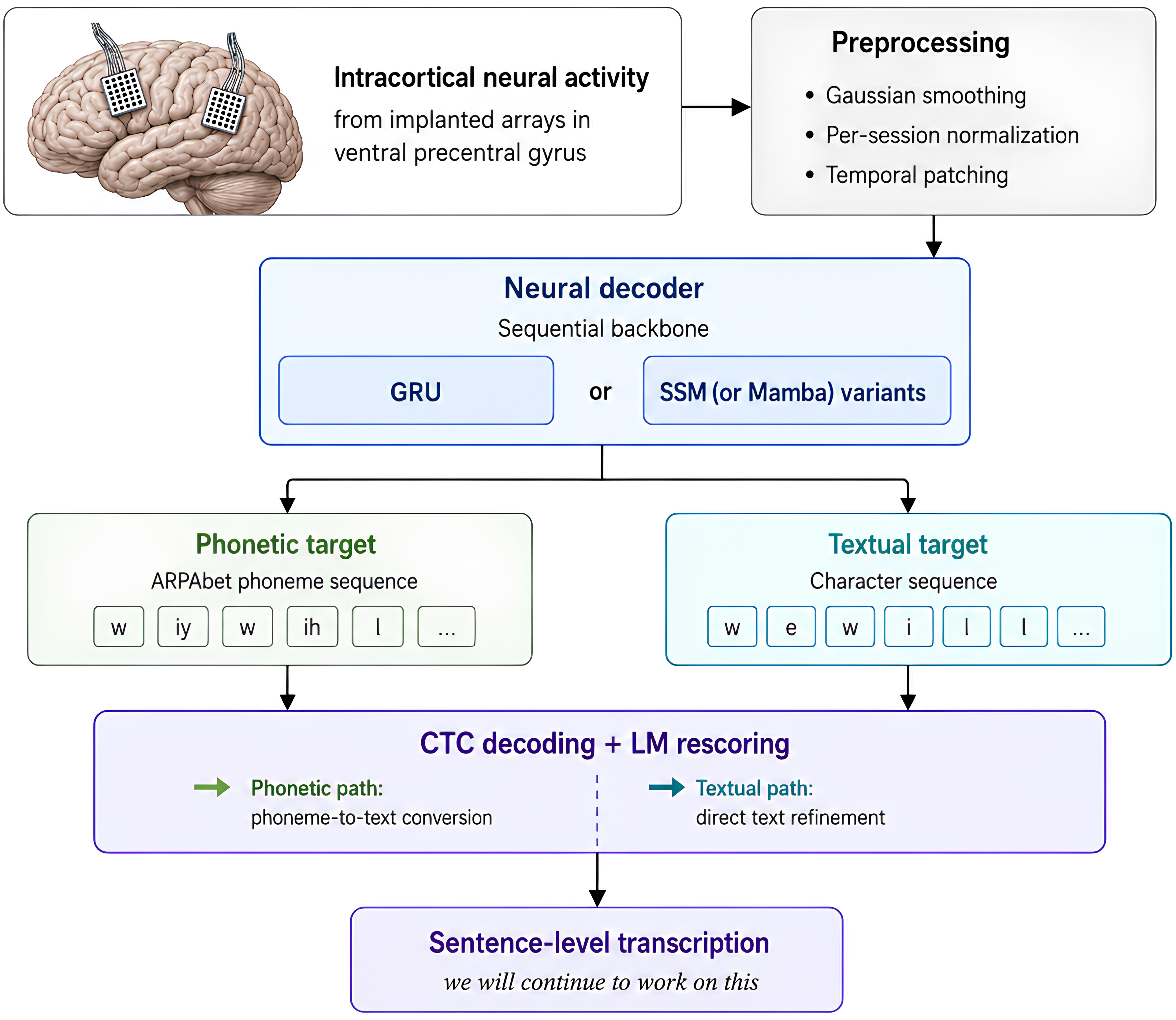}

    \caption{Overview of the experimental pipeline. Intracortical neural activity is  preprocessed and passed to a neural decoder. We study two factors: the sequential backbone (GRU, Mamba variants and ConvMambaGRU) and the target representation (phonetic vs.\ textual). The resulting sequences are decoded with CTC and optionally refined through beam search and language-model rescoring to obtain the final sentence-level transcription.}
    \label{fig:pipeline}
\end{figure}

Fig.~\ref{fig:pipeline} summarizes the pipeline shared by all configurations. A trial of intracortical neural features is first passed through a per-session adaptation layer that normalizes day-to-day distribution shifts, then through temporal patching that shortens the effective sequence. A sequential core (the recurrent baseline or one of the state-space variants) produces a per-step distribution over an output vocabulary, trained end-to-end with a CTC objective~\cite{graves2006ctc}. The two design axes we study sit at the two ends of this core: the \textit{target representation} (phonemes vs.\ characters), which sets the output vocabulary, and the \textit{sequential backbone}. A final linguistic-decoding stage (greedy or beam search) optionally rescored by a language model turns the logits into the output sentence; in the phonetic branch this stage also performs the phoneme-to-text conversion. Keeping every stage fixed except the backbone and the target lets us attribute differences to those two factors.

\subsection{Dataset and signal processing}
\label{sec:data}
We use the public Brain-to-Text '25 benchmark, released with the speech neuroprosthesis of Card et al.~\cite{card2024accurate} and used by the Brain-to-Text benchmark line of work~\cite{willett2024benchmark}. The data come from a single participant with ALS, severe dysarthria and tetraparesis but intact cognition, implanted with four 64-channel microelectrode arrays ($256$ intracortical electrodes) in the left ventral precentral gyrus~\cite{card2024accurate}. Each trial is a prompted speech attempt with an aligned transcript, encoded as a variable-length matrix $X \in \mathbb{R}^{T\times 512}$, where $T$ is the number of $20$~ms time bins and the $512$ features are two measures per electrode (binned threshold-crossing counts and spike-band power). We use the official partitions: $8{,}072$ training and $1{,}426$ validation trials; the $1{,}450$ test trials are excluded as their transcripts are not public, so all comparisons are on validation.

Signal processing follows the baseline pipeline: Gaussian temporal smoothing and per-session normalization, the latter reducing inter-session distribution shift from electrode drift and impedance changes, a well-documented source of non-stationarity in chronic recordings~\cite{simeral2011neural, gallego2017neural} addressed by prior work through recalibration~\cite{jarosiewicz2015virtual} and manifold alignment~\cite{degenhart2020stabilization}. During training only, we apply lightweight SpecAugment-style augmentations~\cite{park2019specaugment} as regularization: per-channel gain, white noise, constant offsets, random-walk noise, and random start crops.

\subsection{Problem formulation}
Given the trial representation $X \in \mathbb{R}^{T\times 512}$ , the decoder emits a sequence of per-frame logits over an output vocabulary of either phonemes or characters. Because there is no explicit alignment between $X$ and the target $y$, we train with CTC~\cite{graves2006ctc}: the probability of $y$ given $X$ is the sum over all alignments $\pi$ that collapse to $y$, $P(y\mid X) = \sum_{\pi \in B^{-1}(y)} P(\pi \mid X)$, and the loss is $\mathcal{L}_{\text{CTC}} = -\log P(y\mid X)$. At inference, CTC collapsing removes blanks and consecutive repeats. Variable-length trials are batched with zero-padding while preserving true lengths, so padding does not affect the loss.

\subsection{Decoder architectures}
All backbones share the pipeline of Section~\ref{sec:procedure} (per-session adaptation, temporal patching, sequential core, linear projection), so performance differences are attributable to the core and the target representation.

\textbf{GRU baseline.} A reproducible recurrent decoder aligned with recent brain-to-text systems: a per-session adaptation layer, temporal patching to shorten the effective sequence, a stack of unidirectional GRU layers~\cite{cho2014gru}, and a final projection.

\textbf{State-space models.} As an alternative core we consider Mamba~\cite{gu2023mamba}, a selective state-space model (SSM). A continuous SSM maps an input signal $x(t)$ to an output $y(t)$ through a latent state $h(t)$:
\begin{align}
  h'(t) &= A\,h(t) + B\,x(t), &
  y(t)  &= C\,h(t),
\end{align}
where $A$ governs the internal state dynamics, $B$ how the input drives the state, and $C$ how the state is read out. Discretizing in time yields a linear recurrence,
\begin{align}
  h_t &= \bar{A}_t\,h_{t-1} + \bar{B}_t\,x_t, &
  y_t &= C_t\,h_t,
\end{align}
which scales linearly with sequence length, unlike the quadratic cost of self-attention, making it attractive for the long, noisy neural sequences here. The key idea of Mamba is \textit{selectivity}: the parameters $(\bar{B}_t, C_t)$ and the discretization step become functions of the current input, so the model can choose which information to keep, update or discard at each step~\cite{gu2023mamba}. This input-dependent gating is well suited to neural recordings, where not every bin is equally informative.

On top of this core we evaluate three increasingly structured variants. \textbf{(i) Mamba} applies residual Mamba blocks directly after patching, isolating the contribution of the SSM core alone~\cite{gu2023mamba, miyazaki2024mamba}. \textbf{(ii) ConvMamba} prepends a residual Conv1D front-end before the Mamba blocks. The motivation, following convolution-augmented SSMs that replace Conformer self-attention with Mamba layers~\cite{hou2024conmamba, zevallos2025mamba}, is that local depthwise convolutions efficiently capture short-range articulatory-like patterns that complement the longer-range dependencies modeled by the SSM, which is particularly useful when training data are scarce~\cite{jiang2024slytherin}. \textbf{(iii) ConvMambaGRU}, our proposed variant, adds a final unidirectional GRU after the Mamba blocks as a temporal-refinement stage before projection, combining local convolutional context, linear-time selective state-space modeling, and recurrent refinement. The temporal patching itself follows patch-based tokenization from vision transformers~\cite{dosovitskiy2021image}, reducing sequence length while preserving local context.

Each backbone is instantiated as a \textit{phonetic} variant (ARPAbet targets, later converted to text by the LM stage) and a \textit{textual} variant (character targets, decoded directly). Table~\ref{tab:arch} summarizes the configurations.

\begin{table}[t]
  \caption{Evaluated backbones, instantiated as phonetic (P) and textual (T) variants. Output classes include the CTC blank.}
  \label{tab:arch}
  \centering
  \setlength{\tabcolsep}{3pt}   % por defecto 6pt; baja a 3-4pt
  \scriptsize
  \begin{tabular}{lcc l}
    \toprule
    \textbf{Backbone} & \textbf{Params (P)} & \textbf{Params (T)} & \textbf{Temporal core} \\
    \midrule
    GRU (baseline) & 44.3M & 44.3M & 5$\times$ GRU \\
    Mamba          & 36.2M & 36.2M & 5$\times$ Mamba \\
    ConvMamba      & 50.6M & 50.6M & Conv1D + 5$\times$ Mamba \\
    ConvMambaGRU   & 50.7M & 54.2M & Conv1D + 5$\times$ Mamba + GRU \\
    \bottomrule
  \end{tabular}
  % \blue{[FILL IN] measure params for Mamba and ConvMamba (P/T columns).}
\end{table}

\subsection{Linguistic decoding and rescoring}
\label{ssec:rescoring}
From the $T\times P$ logit matrix we obtain hypotheses by greedy CTC decoding and by beam search. On the beam-search hypotheses we apply LM rescoring: the final score combines the neural (CTC) score with the LM plausibility, $\text{Score}(y) = \text{Score}_{\text{CTC}}(y) + \alpha\,\text{Score}_{\text{LM}}(y)$, where $\alpha$ weights the linguistic component. In the phonetic variant the LM is essential, as it performs the phoneme-to-text conversion; in the textual variant it refines an already character-level hypothesis. We compare a generalist Transformer LM (GPT-2 variants~\cite{radford2019gpt2}) and a domain-trained $n$-gram LM (KenLM~\cite{heafield2011kenlm}).
For textual decoding, we use CTC beam search with beam width 100 and retain the top 100 hypotheses for LM rescoring. The LM weight $\alpha$ is selected on the validation set for each decoder--LM pair according to the lowest WER.

\section{Experimental setup}

The dataset, splits and signal processing are described in Section~\ref{sec:data}; all results are reported on the $1{,}426$-trial validation set.

\textbf{Metrics.} Phoneme Error Rate (PER) for the phonetic decoder; Character Error Rate (CER) and Word Error Rate (WER) for the final text are used to evaluate the proposed systems.

\textbf{Implementation.} Models are implemented in PyTorch and trained on a single NVIDIA 3060 GPU with 12 GB of VRAM. For reproducibility, configuration files, checkpoints, validation predictions and metric logs are retained for every run. Random seeds are fixed for each experiment. Code and trained-model artifacts will be released with the final version of the paper.

\textbf{Statistical reporting.} For the phonetic backbone comparison, each model is trained with five random seeds, and we report mean and standard deviation of PER across runs. The textual configurations are reported as single-run validation results. For the utterance-level paired analysis, we focus on the GRU baseline and ConvMambaGRU, the best-performing Mamba-based variant in the multi-seed phonetic comparison. Paired differences in PER and WER are computed on the same $1{,}426$ validation trials and assessed with the Wilcoxon signed-rank test~\cite{wilcoxon1945}, following recommended practice for comparing learning systems~\cite{santafe2015evaluation}. We report both two-sided tests and directional tests in the observed direction.

\section{Results}

\subsection{Backbone performance on the phonetic task}
\label{sec:backbone-exploration}
We first explore how the sequential backbones behave on the phonetic task, using validation PER over five independent runs (Table~\ref{tab:ablation}) to identify the strongest SSM variant to carry forward. The GRU baseline obtains the lowest mean PER and is the most stable model. Among the SSM backbones, performance improves as convolutional preprocessing and recurrent refinement are added to the bare Mamba core: Mamba-only and ConvMamba lag clearly behind, MambaGRU is competitive on average but unstable (one diverging run inflates its variance), and ConvMambaGRU is the strongest and most consistent SSM configuration, narrowing the gap to the recurrent baseline. This indicates that selective SSM blocks alone do not improve intracortical phoneme decoding in this data-limited regime, becoming competitive only when combined with both local convolutional context and recurrent refinement, consistent with benchmark observations that recurrent baselines remain hard to beat with limited data~\cite{willett2024benchmark, li2024context}. We therefore adopt ConvMambaGRU as the representative SSM model.

\begin{table}[t]
  \caption{Initial backbone exploration on the phonetic validation task. Values are PER (\%) over five independent runs; the last column reports mean $\pm$ standard deviation across seeds.}
  \label{tab:ablation}
  \centering
  \footnotesize
  \setlength{\tabcolsep}{2.75pt}
  \begin{tabular}{lcccccc}
    \toprule
    \textbf{Backbone} & \textbf{R1} & \textbf{R2} & \textbf{R3} & \textbf{R4} & \textbf{R5} & \textbf{Mean $\pm$ std} \\
    \midrule
    GRU          & 12.62 & 13.15 & 12.79 & 12.55 & 13.68 & \textbf{12.96 $\pm$ 0.47} \\
    Mamba        & 15.37 & 14.93 & 14.99 & 15.03 & 14.57 & 14.98 $\pm$ 0.29 \\
    ConvMamba    & 17.47 & 17.91 & 16.89 & 17.50 & 17.21 & 17.40 $\pm$ 0.38 \\
    MambaGRU     & 14.05 & 13.95 & 14.06 & 13.78 & 22.41 & 15.65 $\pm$ 3.78 \\
    ConvMambaGRU & 13.13 & 13.16 & 13.67 & 13.68 & 13.58 & 13.44 $\pm$ 0.28 \\
    \bottomrule
  \end{tabular}
\end{table}

\subsection{Main comparison}
\begin{table}[t]
    \caption{Main validation results. Phonetic models use the best runs from Table~\ref{tab:ablation} with the official 1-gram OpenWebText baseline LM. Textual models use GPT-2 XL rescoring with the best validation $\alpha$ per decoder. PER/CER are intermediate metrics; WER is the final text metric.}
  \label{tab:main}
  \centering
  \footnotesize
  \begin{tabular}{llcc}
    \toprule
    \textbf{Target} & \textbf{Model} & \textbf{PER/CER (\%)} & \textbf{WER (\%)} \\
    \midrule
    Phonetic & GRU          & 12.62 (PER) & \textbf{21.19} \\
    Phonetic & ConvMambaGRU & 13.13 (PER) & 24.11 \\
    Textual  & GRU          & 13.39 (CER) & 26.28 \\
    Textual  & ConvMambaGRU & 15.29 (CER) & 29.86 \\
    \bottomrule
  \end{tabular}
\end{table}

For the main comparison, final WER for the phonetic systems is computed with the official baseline language model from the Brain-to-Text benchmark pipeline: a 1-gram OpenWebText language model with silence-token handling. For the textual systems, final WER is computed after character-level beam search and GPT-2 XL rescoring, using the best validation weight for each decoder ($\alpha=1.5$ for GRU and $\alpha=2.0$ for ConvMambaGRU; Table~\ref{tab:lm}).

We compare ConvMambaGRU, the strongest SSM backbone, against the GRU baseline across both target representations (Table~\ref{tab:main}); phonetic values correspond to the best run from Table~\ref{tab:ablation}, textual models are single-run results. The GRU is the strongest decoder in both representations, and the two-stage phonetic models outperform their textual counterparts ($21.19\%$ vs.\ $26.28\%$ WER). ConvMambaGRU is competitive on PER but does not surpass the GRU on final WER. We speculate that the phonetic advantage arises because the neural signal, originating in speech-motor cortex, encodes articulatory structure more directly than orthographic characters, placing a phonetic target closer to the information present in the recordings. The textual variant nonetheless retains practical appeal: decoding characters directly removes the need for a pronunciation lexicon and phoneme-to-text conversion, although in this study its best performance still requires GPT-2 XL rescoring. Finally, small intermediate differences propagate non-linearly: because the LM resolves hypotheses at the word level, a sub-1-point PER change can shift lexical and word-boundary decisions into a $\sim$3-point WER gap, confirming that final performance is the joint product of the neural decoder and the linguistic stage~\cite{willett2024benchmark, li2024context}.

\subsection{Effect of the language model}
We next examine the linguistic decoding stage, sweeping the rescoring weight $\alpha$ described in Section~\ref{ssec:rescoring} for both decoders and several language models on the textual target. Table~\ref{tab:lm} reports, for each neural decoder and LM, the configuration with the lowest validation WER. The domain $n$-gram (KenLM) and the generalist GPT-2 variants trade off differently; rescoring consistently improves over greedy decoding, but cannot fully compensate for a weaker neural decoder.

\begin{table}[t]
  \caption{Language-model rescoring for the textual target on the validation set. For each neural decoder and language model, $\alpha$ is selected according to the lowest validation WER.}
  \label{tab:lm}
  \centering
  \footnotesize
  \setlength{\tabcolsep}{4pt}
  \begin{tabular}{llccc}
    \toprule
    \textbf{Model} & \textbf{LM} & \textbf{CER (\%)} & \textbf{WER (\%)} & $\boldsymbol{\alpha}$ \\
    \midrule
    GRU & GPT-2 XL     & 13.39 & 26.28 & 1.5 \\
    GRU & GPT-2        & 13.51 & 26.44 & 2.0 \\
    GRU & KenLM 5-gram & 14.19 & 27.80 & 5.0 \\
    GRU & KenLM 3-gram & 14.20 & 27.88 & 4.5 \\
    \midrule
    ConvMambaGRU & GPT-2 XL     & 15.29 & 29.86 & 2.0 \\
    ConvMambaGRU & KenLM 3-gram & 16.13 & 30.94 & 7.5 \\
    ConvMambaGRU & KenLM 5-gram & 16.13 & 30.95 & 7.5 \\
    ConvMambaGRU & GPT-2        & 15.85 & 31.10 & 0.8 \\
    \bottomrule
  \end{tabular}
\end{table}

\subsection{Error analysis}
Errors grow with sentence length in both variants, and there is appreciable inter-session variability, consistent with the exploratory analysis of signal stability. Crucially, the two representations fail differently, and the target determines how errors become visible.

In the phonetic variant, substitutions are not uniform across the inventory. Table~\ref{tab:phoneme_conf} lists the most frequent phoneme substitutions, in both ARPAbet and IPA, aggregated over the five GRU phonetic validation runs. The dominant confusions are reciprocal and phonetically plausible: the most frequent pair is D \textipa{/d/}$\rightarrow$T \textipa{/t/}, followed by its reverse, and the same voicing contrast appears for Z \textipa{/z/}$\rightarrow$S \textipa{/s/} and S \textipa{/s/}$\rightarrow$Z \textipa{/z/}. Vowel confusions between neighbouring qualities are also prominent (e.g.\ AH \textipa{/@/}$\leftrightarrow$IH \textipa{/I/}, IH \textipa{/I/}$\rightarrow$EH \textipa{/E/}, IY \textipa{/i/}$\rightarrow$IH \textipa{/I/}). These patterns indicate systematic difficulty discriminating articulatorily and acoustically close phones rather than random sequence-level errors.

\begin{table}[t]
  \caption{Most frequent phoneme substitutions for the phonetic variant on the validation set. Counts are aggregated over the five GRU phonetic validation runs.}
  \label{tab:phoneme_conf}
  \centering
  \footnotesize
  \begin{tabular}{llr}
    \toprule
    \textbf{Reference} & \textbf{Hypothesis} & \textbf{Count} \\
    \midrule
    D \textipa{/d/}  & T \textipa{/t/}  & 631 \\
    Z \textipa{/z/}  & S \textipa{/s/}  & 354 \\
    AH \textipa{/@/} & IH \textipa{/I/} & 349 \\
    T \textipa{/t/}  & D \textipa{/d/}  & 339 \\
    IH \textipa{/I/} & AH \textipa{/@/} & 272 \\
    S \textipa{/s/}  & Z \textipa{/z/}  & 267 \\
    IH \textipa{/I/} & EH \textipa{/E/} & 239 \\
    IY \textipa{/i/} & IH \textipa{/I/} & 233 \\
    \bottomrule
  \end{tabular}
\end{table}

In the textual variant, the most frequent errors concentrate on short, high-frequency words (Table~\ref{tab:lex_conf}), mainly function-word and near-homophone substitutions such as \textit{the}$\rightarrow$\textit{they}, \textit{do}$\rightarrow$\textit{to}, \textit{too}$\rightarrow$\textit{to}, and \textit{their}$\rightarrow$\textit{there}. Even after beam search with LM rescoring, the decoder can still confuse brief lexical items whose acoustic, orthographic or contextual evidence is weak. This contrasts with the phonetic analysis, where errors surface as local phone substitutions, and is complementary to a recent character-level Conformer study on the same data, which reports dominant character-level errors involving the space token~\cite{khanday2026endtoend}. Together, the two analyses show that the target representation governs how decoding errors become visible: phonetic targets expose articulatory and acoustic confusions, whereas textual targets expose lexical and segmentation-sensitive errors.

\begin{table}[t]
  \caption{Most frequent word-level confusions for the textual variant on the validation set, using beam search with language-model rescoring.}
  \label{tab:lex_conf}
  \centering
  \footnotesize
  \begin{tabular}{llr}
    \toprule
    \textbf{Reference} & \textbf{Hypothesis} & \textbf{Count} \\
    \midrule
    the   & they  & 9 \\
    do    & to    & 8 \\
    their & there & 7 \\
    too   & to    & 6 \\
    has   & as    & 5 \\
    in    & it    & 5 \\
    \bottomrule
  \end{tabular}
\end{table}

\section{Conclusion}
We presented a controlled study of intracortical brain-to-text decoding crossing the target representation (phonetic vs.\ textual) with the decoder architecture (recurrent vs.\ a selective state-space hybrid) under one reproducible CTC pipeline. The recurrent baseline remains the strongest decoder, and the phonetic two-stage variant gives the lowest WER. In particular, the GRU neural decoder obtains lower error than ConvMambaGRU in PER (12.62\% vs.\ 13.13\%, a difference of -0.51 percentage points; Wilcoxon two-sided $p=0.023$; directional $p=0.0115$) and in WER (21.19\% vs.\ 24.11\%, a difference of -2.92 percentage points; Wilcoxon two-sided $p=3.7\times10^{-10}$; directional $p=1.85\times10^{-10}$). Final performance is therefore the joint product of the neural decoder and the linguistic stage, and the dominant error modes are representation-specific. Future work will explore three directions aligned with recent research on low-resource intracortical decoding~\cite{willett2024benchmark, li2024context}: richer target representations, stronger training and rescoring strategies, and more task-specific adaptation of SSM architectures to neural signals, rather than simply substituting the recurrent backbone.

\section{Acknowledgments}
% \red{NOTE (manual): remove this section in the double-blind review version; fill in grant IDs for camera-ready.}
\ifcameraready
    This work was supported by grants PID2022-141378OB-C22 and AIA2025-163317-C32 funded by MICIU/AEI/10.13039/501100011033 and ERDF/EU.
\else
    Acknowledgments withheld for double-blind review.
\fi
\ %

\section{Generative AI Use Disclosure}
During the preparation of this work, the authors used generative AI tools for language editing and to assist with code development. These tools were not used to generate scientific content, design the methodology, or interpret the results. After using these tools, the authors reviewed and edited the content as needed and take full responsibility for the content of the publication.

\bibliographystyle{IEEEtran}
\bibliography{mybib}

@article{silva2024speech,
  author  = {Silva, Alexander B. and Littlejohn, Kaylo T. and Liu, Jessie R. and Moses, David A. and Chang, Edward F.},
  title   = {The speech neuroprosthesis},
  journal = {Nature Reviews Neuroscience},
  volume  = {25},
  number  = {7},
  pages   = {473--492},
  year    = {2024}
}

@article{card2024accurate,
  author  = {Card, Nicholas S. and Wairagkar, Maitreyee and Iacobacci, Carrina and Bhatt, Payal and Singer-Clark, Tyler and Willett, Francis R. and Ames, Kelly C. and Liu, Jana and Rezaii, Paymon and Hochberg, Leigh R. and Henderson, Jaimie M. and Shenoy, Krishna V. and Brandman, David M.},
  title   = {An accurate and rapidly calibrating speech neuroprosthesis},
  journal = {New England Journal of Medicine},
  volume  = {391},
  number  = {7},
  pages   = {609--618},
  year    = {2024}
}

@article{willett2023high,
  author  = {Willett, Francis R. and Kunz, Erin M. and Fan, Chaofei and Avansino, Donald T. and Wilson, Guy H. and Choi, Eun Young and Kamdar, Foram and Hochberg, Leigh R. and Henderson, Jaimie M. and Shenoy, Krishna V.},
  title   = {A high-performance speech neuroprosthesis},
  journal = {Nature},
  volume  = {620},
  number  = {7976},
  pages   = {1031--1036},
  year    = {2023}
}

@article{moses2021neuroprosthesis,
  author  = {Moses, David A. and Metzger, Sean L. and Liu, Jessie R. and Anumanchipalli, Gopala K. and Makin, Joseph G. and Sun, Pengfei F. and Chartier, Josh and Dougherty, Maximilian E. and Liu, Patricia M. and Abrams, Gary M. and Tu-Chan, Adelyn and Ganguly, Karunesh and Chang, Edward F.},
  title   = {Neuroprosthesis for decoding speech in a paralyzed person with anarthria},
  journal = {New England Journal of Medicine},
  volume  = {385},
  number  = {3},
  pages   = {217--227},
  year    = {2021}
}

@article{willett2021handwriting,
  author  = {Willett, Francis R. and Avansino, Donald T. and Hochberg, Leigh R. and Henderson, Jaimie M. and Shenoy, Krishna V.},
  title   = {High-performance brain-to-text communication via handwriting},
  journal = {Nature},
  volume  = {593},
  number  = {7858},
  pages   = {249--254},
  year    = {2021}
}

@article{metzger2023avatar,
  author  = {Metzger, Sean L. and Littlejohn, Kaylo T. and Silva, Alexander B. and Moses, David A. and Seaton, Margaret P. and Wang, Ran and Dougherty, Maximilian E. and Liu, Jessie R. and Wu, Peter and Berger, Michael A. and Zhuravleva, Inga and Tu-Chan, Adelyn and Ganguly, Karunesh and Anumanchipalli, Gopala K. and Chang, Edward F.},
  title   = {A high-performance neuroprosthesis for speech decoding and avatar control},
  journal = {Nature},
  volume  = {620},
  number  = {7976},
  pages   = {1037--1046},
  year    = {2023}
}

@article{metzger2022spelling,
  author  = {Metzger, Sean L. and Liu, Jessie R. and Moses, David A. and Dougherty, Maximilian E. and Seaton, Margaret P. and Littlejohn, Kaylo T. and Chartier, Josh and Anumanchipalli, Gopala K. and Tu-Chan, Adelyn and Ganguly, Karunesh and Chang, Edward F.},
  title   = {Generalizable spelling using a speech neuroprosthesis in an individual with severe limb and vocal paralysis},
  journal = {Nature Communications},
  volume  = {13},
  pages   = {6510},
  year    = {2022}
}

@article{makin2020machine,
  author  = {Makin, Joseph G. and Moses, David A. and Chang, Edward F.},
  title   = {Machine translation of cortical activity to text with an encoder--decoder framework},
  journal = {Nature Neuroscience},
  volume  = {23},
  number  = {4},
  pages   = {575--582},
  year    = {2020}
}

@article{anumanchipalli2019synthesis,
  author  = {Anumanchipalli, Gopala K. and Chartier, Josh and Chang, Edward F.},
  title   = {Speech synthesis from neural decoding of spoken sentences},
  journal = {Nature},
  volume  = {568},
  number  = {7753},
  pages   = {493--498},
  year    = {2019}
}

@article{wairagkar2025voice,
  author  = {Wairagkar, Maitreyee and Card, Nicholas S. and Singer-Clark, Tyler and Hou, Xianda and Iacobacci, Carrina and Hochberg, Leigh R. and Brandman, David M. and Stavisky, Sergey D.},
  title   = {An instantaneous voice-synthesis neuroprosthesis},
  journal = {Nature},
  volume  = {644},
  pages   = {145--152},
  year    = {2025}
}

@article{wolpaw2002bci,
  author  = {Wolpaw, Jonathan R. and Birbaumer, Niels and McFarland, Dennis J. and Pfurtscheller, Gert and Vaughan, Theresa M.},
  title   = {Brain-computer interfaces for communication and control},
  journal = {Clinical Neurophysiology},
  volume  = {113},
  number  = {6},
  pages   = {767--791},
  year    = {2002}
}

@article{hickok2007cortical,
  author  = {Hickok, Gregory and Poeppel, David},
  title   = {The cortical organization of speech processing},
  journal = {Nature Reviews Neuroscience},
  volume  = {8},
  number  = {5},
  pages   = {393--402},
  year    = {2007}
}

@inproceedings{gulati2020conformer,
  author    = {Gulati, Anmol and Qin, James and Chiu, Chung-Cheng and Parmar, Niki and Zhang, Yu and Yu, Jiahui and Han, Wei and Wang, Shibo and Zhang, Zhengdong and Wu, Yonghui and Pang, Ruoming},
  title     = {Conformer: Convolution-augmented transformer for speech recognition},
  booktitle = {Proc. Interspeech},
  pages     = {5036--5040},
  year      = {2020}
}

@inproceedings{vaswani2017attention,
  author    = {Vaswani, Ashish and Shazeer, Noam and Parmar, Niki and Uszkoreit, Jakob and Jones, Llion and Gomez, Aidan N. and Kaiser, Lukasz and Polosukhin, Illia},
  title     = {Attention is all you need},
  booktitle = {Advances in Neural Information Processing Systems},
  volume    = {30},
  year      = {2017}
}

@inproceedings{graves2006ctc,
  author    = {Graves, Alex and Fern{\'a}ndez, Santiago and Gomez, Faustino and Schmidhuber, J{\"u}rgen},
  title     = {Connectionist temporal classification: Labelling unsegmented sequence data with recurrent neural networks},
  booktitle = {Proc. International Conference on Machine Learning (ICML)},
  pages     = {369--376},
  year      = {2006}
}

@inproceedings{cho2014gru,
  author    = {Cho, Kyunghyun and van Merri{\"e}nboer, Bart and Gulcehre, Caglar and Bahdanau, Dzmitry and Bougares, Fethi and Schwenk, Holger and Bengio, Yoshua},
  title     = {Learning phrase representations using {RNN} encoder--decoder for statistical machine translation},
  booktitle = {Proc. Conference on Empirical Methods in Natural Language Processing (EMNLP)},
  pages     = {1724--1734},
  year      = {2014}
}

@inproceedings{zevallos2025mamba,
  author    = {Zevallos, Rodolfo and Cortada-Garcia, Marc and Solito, Sara and Mena, Carlos and Peiro-Lilja, Alex and Hernando, Javier},
  title     = {Assessing the performance and efficiency of {Mamba} {ASR} in low-resource scenarios},
  booktitle = {Proc. Interspeech},
  pages     = {5198--5202},
  year      = {2025}
}

@inproceedings{heafield2011kenlm,
  author    = {Heafield, Kenneth},
  title     = {{KenLM}: Faster and smaller language model queries},
  booktitle = {Proc. Sixth Workshop on Statistical Machine Translation (WMT)},
  pages     = {187--197},
  year      = {2011}
}

@inproceedings{park2019specaugment,
  author    = {Park, Daniel S. and Chan, William and Zhang, Yu and Chiu, Chung-Cheng and Zoph, Barret and Cubuk, Ekin D. and Le, Quoc V.},
  title     = {{SpecAugment}: A simple data augmentation method for automatic speech recognition},
  booktitle = {Proc. Interspeech},
  pages     = {2613--2617},
  year      = {2019}
}

@article{radford2023whisper,
  author  = {Radford, Alec and Kim, Jong Wook and Xu, Tao and Brockman, Greg and McLeavey, Christine and Sutskever, Ilya},
  title   = {Robust speech recognition via large-scale weak supervision},
  journal = {Proc. International Conference on Machine Learning (ICML)},
  pages   = {28492--28518},
  year    = {2023}
}

@inproceedings{baevski2020wav2vec,
  author    = {Baevski, Alexei and Zhou, Yuhao and Mohamed, Abdelrahman and Auli, Michael},
  title     = {wav2vec 2.0: A framework for self-supervised learning of speech representations},
  booktitle = {Advances in Neural Information Processing Systems},
  volume    = {33},
  pages     = {12449--12460},
  year      = {2020}
}

@techreport{radford2019gpt2,
  author      = {Radford, Alec and Wu, Jeffrey and Child, Rewon and Luan, David and Amodei, Dario and Sutskever, Ilya},
  title       = {Language models are unsupervised multitask learners},
  institution = {OpenAI},
  year        = {2019}
}

@article{degenhart2020stabilization,
  author  = {Degenhart, Alan D. and Bishop, William E. and Oby, Emily R. and Tyler-Kabara, Elizabeth C. and Chase, Steven M. and Batista, Aaron P. and Yu, Byron M.},
  title   = {Stabilization of a brain-computer interface via the alignment of low-dimensional spaces of neural activity},
  journal = {Nature Biomedical Engineering},
  volume  = {4},
  number  = {7},
  pages   = {672--685},
  year    = {2020}
}

@article{jarosiewicz2015virtual,
  author  = {Jarosiewicz, Beata and Sarma, Anish A. and Bacher, Daniel and Masse, Nicolas Y. and Simeral, John D. and Sorice, Brittany and Oakley, Erin M. and Blabe, Christine and Pandarinath, Chethan and Gilja, Vikash and Cash, Sydney S. and Eskandar, Emad N. and Friehs, Gerhard and Henderson, Jaimie M. and Shenoy, Krishna V. and Donoghue, John P. and Hochberg, Leigh R.},
  title   = {Virtual typing by people with tetraplegia using a self-calibrating intracortical brain-computer interface},
  journal = {Science Translational Medicine},
  volume  = {7},
  number  = {313},
  pages   = {313ra179},
  year    = {2015}
}

@article{wilcoxon1945,
  author  = {Wilcoxon, Frank},
  title   = {Individual comparisons by ranking methods},
  journal = {Biometrics Bulletin},
  volume  = {1},
  number  = {6},
  pages   = {80--83},
  year    = {1945}
}

@article{santafe2015evaluation,
  author  = {Santaf{\'e}, Guzm{\'a}n and Inza, I{\~n}aki and Lozano, Jose A.},
  title   = {Dealing with the evaluation of supervised classification algorithms},
  journal = {Artificial Intelligence Review},
  volume  = {44},
  number  = {4},
  pages   = {467--508},
  year    = {2015}
}

@article{gu2023mamba,
  author  = {Gu, Albert and Dao, Tri},
  title   = {Mamba: Linear-time sequence modeling with selective state spaces},
  journal = {arXiv preprint arXiv:2312.00752},
  year    = {2023}
}

@article{willett2024benchmark,
  author  = {Willett, Francis R. and Li, Jingyuan and Le, Trung and Fan, Chaofei and Chen, Mingfei and Shlizerman, Eli},
  title   = {Brain-to-Text Benchmark '24: Lessons learned},
  journal = {arXiv preprint arXiv:2412.17227},
  year    = {2024}
}

@article{li2024context,
  author  = {Li, Jingyuan and Le, Trung and Fan, Chaofei and Chen, Mingfei and Shlizerman, Eli},
  title   = {Brain-to-Text decoding with context-aware neural representations and large language models},
  journal = {arXiv preprint arXiv:2411.10657},
  year    = {2024}
}

@article{khanday2026endtoend,
author = {Khanday, Owais Mujtaba and Gonzalez-Lopez, Jose A. and Ouellet, Marc and Galdon, Alberto and Olivares Granados, Gonzalo},
title = {End-to-End Intracortical Speech Decoding from Neural Activity},
journal = {arXiv preprint arXiv:2605.24313},
year = {2026}
}

@article{chang2024restoring,
  author  = {Chang, Edward F.},
  title   = {Brain-computer interfaces for restoring communication},
  journal = {New England Journal of Medicine},
  volume  = {391},
  number  = {7},
  pages   = {654--657},
  year    = {2024}
}

@article{herff2015braintotext,
  author  = {Herff, Christian and Heger, Dominic and de Pesters, Adriana and Telaar, Dominic and Brunner, Peter and Schalk, Gerwin and Schultz, Tanja},
  title   = {Brain-to-text: Decoding spoken phrases from phone representations in the brain},
  journal = {Frontiers in Neuroscience},
  volume  = {9},
  pages   = {217},
  year    = {2015}
}

@article{angrick2019synthesis,
  author  = {Angrick, Miguel and Herff, Christian and Mugler, Emily and Tate, Matthew C. and Slutzky, Marc W. and Krusienski, Dean J. and Schultz, Tanja},
  title   = {Speech synthesis from {ECoG} using densely connected {3D} convolutional neural networks},
  journal = {Journal of Neural Engineering},
  volume  = {16},
  number  = {3},
  pages   = {036019},
  year    = {2019}
}

@article{martin2014decoding,
  author  = {Martin, Stephanie and Brunner, Peter and Holdgraf, Chris and Heinze, Hans-Jochen and Crone, Nathan E. and Rieger, Jochem and Schalk, Gerwin and Knight, Robert T. and Pasley, Brian N.},
  title   = {Decoding spectrotemporal features of overt and covert speech from the human cortex},
  journal = {Frontiers in Neuroengineering},
  volume  = {7},
  pages   = {14},
  year    = {2014}
}

@article{brumberg2010bci,
  author  = {Brumberg, Jonathan S. and Nieto-Castanon, Alfonso and Kennedy, Philip R. and Guenther, Frank H.},
  title   = {Brain-computer interfaces for speech communication},
  journal = {Speech Communication},
  volume  = {52},
  number  = {4},
  pages   = {367--379},
  year    = {2010}
}

@article{tang2023semantic,
  author  = {Tang, Jerry and LeBel, Amanda and Jain, Shailee and Huth, Alexander G.},
  title   = {Semantic reconstruction of continuous language from non-invasive brain recordings},
  journal = {Nature Neuroscience},
  volume  = {26},
  number  = {5},
  pages   = {858--866},
  year    = {2023}
}

@article{hochberg2012reach,
  author  = {Hochberg, Leigh R. and Bacher, Daniel and Jarosiewicz, Beata and Masse, Nicolas Y. and Simeral, John D. and Vogel, Joern and Haddadin, Sami and Liu, Jie and Cash, Sydney S. and van der Smagt, Patrick and Donoghue, John P.},
  title   = {Reach and grasp by people with tetraplegia using a neurally controlled robotic arm},
  journal = {Nature},
  volume  = {485},
  number  = {7398},
  pages   = {372--375},
  year    = {2012}
}

@article{pandarinath2017high,
  author  = {Pandarinath, Chethan and Nuyujukian, Paul and Blabe, Christine H. and Sorice, Brittany L. and Saab, Jad and Willett, Francis R. and Hochberg, Leigh R. and Shenoy, Krishna V. and Henderson, Jaimie M.},
  title   = {High performance communication by people with paralysis using an intracortical brain-computer interface},
  journal = {eLife},
  volume  = {6},
  pages   = {e18554},
  year    = {2017}
}

@article{vansteensel2016fully,
  author  = {Vansteensel, Mariska J. and Pels, Elmar G. M. and Bleichner, Martin G. and Branco, Mariana P. and Denison, Timothy and Freudenburg, Zachary V. and Gosselaar, Peter and Leinders, Sacha and Ottens, Thomas H. and van den Boom, Max A. and van Rijen, Peter C. and Aarnoutse, Erik J. and Ramsey, Nick F.},
  title   = {Fully implanted brain-computer interface in a locked-in patient with {ALS}},
  journal = {New England Journal of Medicine},
  volume  = {375},
  number  = {21},
  pages   = {2060--2066},
  year    = {2016}
}

@article{simeral2011neural,
  author  = {Simeral, John D. and Kim, Sung-Phil and Black, Michael J. and Donoghue, John P. and Hochberg, Leigh R.},
  title   = {Neural control of cursor trajectory and click by a human with tetraplegia 1000 days after implant of an intracortical microelectrode array},
  journal = {Journal of Neural Engineering},
  volume  = {8},
  number  = {2},
  pages   = {025027},
  year    = {2011}
}

@article{gallego2017neural,
  author  = {Gallego, Juan A. and Perich, Matthew G. and Miller, Lee E. and Solla, Sara A.},
  title   = {Neural manifolds for the control of movement},
  journal = {Neuron},
  volume  = {94},
  number  = {5},
  pages   = {978--984},
  year    = {2017}
}

@article{pulvermuller2018neural,
  author  = {Pulverm{\"u}ller, Friedemann},
  title   = {Neural reuse of action perception circuits for language, concepts and communication},
  journal = {Progress in Neurobiology},
  volume  = {160},
  pages   = {1--44},
  year    = {2018}
}

@inproceedings{hou2024conmamba,
  author    = {Hou, Haoxiang and others},
  title     = {{ConMamba}: A convolution-augmented {Mamba} encoder model for efficient end-to-end {ASR} systems},
  booktitle = {Proc.\ IEEE International Conference on Signal, Information and Data Processing (ICSIDP)},
  year      = {2024}
}

@inproceedings{miyazaki2024mamba,
  author    = {Miyazaki, Koichi and Masuyama, Yoshiki and Murata, Masato},
  title     = {Exploring the capability of {Mamba} in speech applications},
  booktitle = {Proc. Interspeech},
  pages     = {176--180},
  year      = {2024}
}

@article{jiang2024slytherin,
  author  = {Jiang, Xilin and Li, Yinghao Aaron and Florea, Adrian Nicolas and Han, Cong and Mesgarani, Nima},
  title   = {Speech {Slytherin}: Examining the performance and efficiency of {Mamba} for speech separation, recognition, and synthesis},
  journal = {arXiv preprint arXiv:2407.09732},
  year    = {2024}
}

@inproceedings{burchi2021efficient,
  author    = {Burchi, Maxime and Vielzeuf, Valentin},
  title     = {Efficient {Conformer}: Progressive downsampling and grouped attention for automatic speech recognition},
  booktitle = {Proc.\ IEEE Automatic Speech Recognition and Understanding Workshop (ASRU)},
  pages     = {8--15},
  year      = {2021}
}

@inproceedings{peng2022branchformer,
  author    = {Peng, Yifan and Dalmia, Siddharth and Lane, Ian and Watanabe, Shinji},
  title     = {{Branchformer}: Parallel {MLP}-attention architectures to capture local and global context for speech recognition and understanding},
  booktitle = {Proc. International Conference on Machine Learning (ICML)},
  pages     = {17627--17643},
  year      = {2022}
}

@inproceedings{dosovitskiy2021image,
  author    = {Dosovitskiy, Alexey and Beyer, Lucas and Kolesnikov, Alexander and Weissenborn, Dirk and Zhai, Xiaohua and Unterthiner, Thomas and Dehghani, Mostafa and Minderer, Matthias and Heigold, Georg and Gelly, Sylvain and Uszkoreit, Jakob and Houlsby, Neil},
  title     = {An image is worth 16x16 words: Transformers for image recognition at scale},
  booktitle = {Proc. International Conference on Learning Representations (ICLR)},
  year      = {2021}
}

@inproceedings{chan2016listen,
  author    = {Chan, William and Jaitly, Navdeep and Le, Quoc and Vinyals, Oriol},
  title     = {Listen, attend and spell: A neural network for large vocabulary conversational speech recognition},
  booktitle = {Proc.\ IEEE International Conference on Acoustics, Speech and Signal Processing (ICASSP)},
  pages     = {4960--4964},
  year      = {2016}
}

@article{defossez2023decoding,
  author  = {D{\'e}fossez, Alexandre and Caucheteux, Charlotte and Rapin, J{\'e}r{\'e}my and Kabeli, Ori and King, Jean-R{\'e}mi},
  title   = {Decoding speech perception from non-invasive brain recordings},
  journal = {Nature Machine Intelligence},
  volume  = {5},
  number  = {10},
  pages   = {1097--1107},
  year    = {2023}
}

\end{document}